\documentclass[10pt,twocolumn,letterpaper]{article}

\usepackage{iccv}
\usepackage{times}
\usepackage{epsfig}
\usepackage{graphicx}
\usepackage{amsmath}
\usepackage{amssymb}
\usepackage{iccv}
\usepackage{times}
\usepackage{epsfig}
\usepackage{graphicx}
\usepackage{amsmath}
\usepackage{amssymb}
\usepackage{booktabs}
\usepackage{tabularx}
\usepackage{multirow}
\usepackage{listings}
\usepackage{cite}

\usepackage[breaklinks=true,bookmarks=false]{hyperref}

\iccvfinalcopy % *** Uncomment this line for the final submission

\def\iccvPaperID{****} % *** Enter the ICCV Paper ID here

\ificcvfinal\fi

\begin{document}

%%%%%%%%% TITLE
\title{Understanding Video Scenes through Text: Insights from Text-based Video Question Answering}

\author{\quad Soumya Jahagirdar$^{1}$  \qquad Minesh Mathew$^{2}$ \qquad Dimosthenis Karatzas$^{3}$ \qquad C. V. Jawahar$^{1}$ \\
{\tt\footnotesize soumya.jahagirdar@research.iiit.ac.in} \enspace \enspace {\tt\footnotesize minesh@wadhwaniai.org} \enspace \enspace {\tt\footnotesize dimos@cvc.uab.es}  \enspace\enspace {\tt\footnotesize jawahar@iiit.ac.in} \\
$^{1}$ CVIT, IIIT Hyderabad, India \enspace\enspace $^{2}$ Wadhwani AI \enspace\enspace $^{3}$ Computer Vision Center, UAB, Spain}

% \and

\maketitle
\ificcvfinal\thispagestyle{empty}\fi

%%%%%%%%% ABSTRACT
\begin{abstract}
   Researchers have extensively studied the field of vision and language, discovering that both visual and textual content is crucial for understanding scenes effectively. Particularly, comprehending text in videos holds great significance, requiring both scene text understanding and temporal reasoning. This paper focuses on exploring two recently introduced datasets, NewsVideoQA and M4-ViteVQA, which aim to address video question answering based on textual content. The NewsVideoQA dataset contains question-answer pairs related to the text in news videos, while M4-ViteVQA comprises question-answer pairs from diverse categories like vlogging, traveling, and shopping. We provide an analysis of the formulation of these datasets on various levels, exploring the degree of visual understanding and multi-frame comprehension required for answering the questions. Additionally, the study includes experimentation with BERT-QA, a text-only model, which demonstrates comparable performance to the original methods on both datasets, indicating the shortcomings in the formulation of these datasets. Furthermore, we also look into the domain adaptation aspect by examining the effectiveness of training on M4-ViteVQA and evaluating on NewsVideoQA and vice-versa, thereby shedding light on the challenges and potential benefits of out-of-domain training. 
\end{abstract}

%%%%%%%%% BODY TEXT
%%%%%%%%% BODY TEXT
\section{Introduction}

\begin{figure} 
    \centering
    \includegraphics[width=1\linewidth]{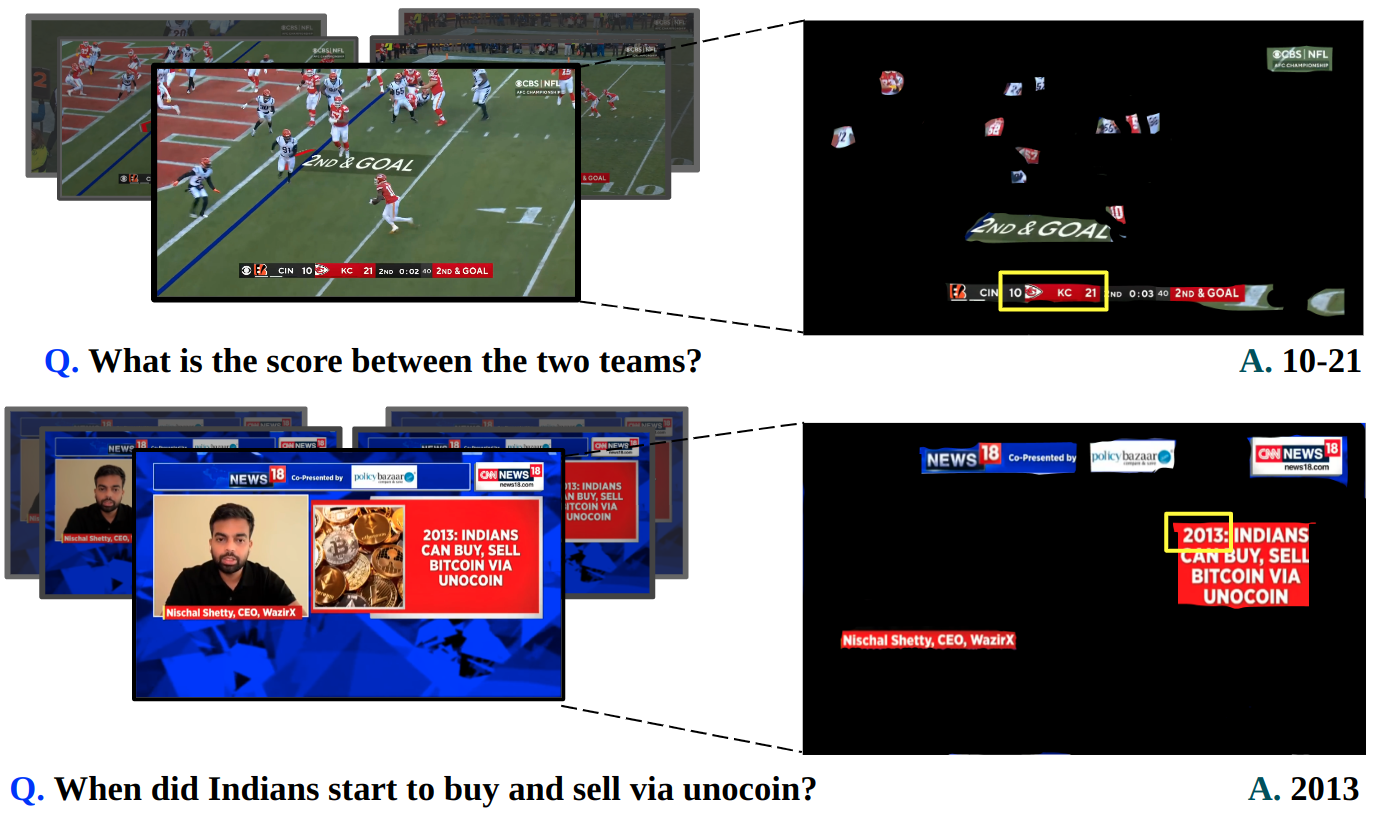}
    \caption{ Example illustrating two major concerns of existing text-based VideoQA datasets \cite{vitevqa_neurips_2022, newsvideoqa_wacv_2023}. Both examples showcase that only \textbf{textual information} from a \textbf{single frame }is sufficient to obtain answers to the questions.
    }
    \label{fig:task}
\end{figure}

Multimodal understanding, specifically, VideoQA is a challenging yet crucial problem that involves multimodal and temporal reasoning. Researchers have developed various datasets and methods to facilitate research in this field \cite{videoqa_gradually_refined_attn_acmm_2017, msrvtt_cvpr_2016, tgif_cvpr_2017, activitnetqa_aaai_2019, moviesqa_cvpr_2016, tvqa_emnlp_2018, howto100m_iccv_2019}. Xu et al. \cite{msrvtt_cvpr_2016}  and Yu et al. \cite{activitnetqa_aaai_2019} propose datasets that contain questions about the events happening in the video but disregard the text. However, works such as Lei et al. \cite{tvqa_emnlp_2018} and Tapaswi et al. \cite{moviesqa_cvpr_2016} have introduced datasets that use both visual and subtitle information to understand the story. However, these existing datasets lacked the ability to handle questions that require reading text in videos. As textual content in man-made environments carries significant semantic information, the need for visual question-answering datasets that involve reading text became evident. These text-based systems have great potential for real-life scenarios, particularly for visually-impaired users and the development of assistive devices. While previous works have explored single-image scene-text and document images \cite{stvqa_iccv_2019, textvqa_cvpr_2019, docvqa_wacv_2021, infographicvqa_wacv_2022, ocrvqa_icdar_2019, vizwiz_cvpr_2018, textvqg_icdar_2021} there has been limited exploration of works that require extracting information from text present in videos. Recently, Hegde et al. \cite{making_v_in_textvqa_matter} shed light on the bias aspects of TextVQA datasets. Recently, there have been multiple datasets \cite{newsvideoqa_wacv_2023, vitevqa_neurips_2022, roadtextvqa_icdar_2023} that deal with a new line of research wherein the datasets require models to read and understand the text present in the videos to answer the questions. Jahagirdar et al. \cite{newsvideoqa_wacv_2023} proposed a dataset: NewsVideoQA that contains question-answer pairs framed on news videos from multiple news channels, and these questions are formulated such that answering the questions requires an understanding of the embedded text i.e. text occurring in the news videos. Similarly, Zhao et al. \cite{vitevqa_neurips_2022} introduced a dataset that contains videos from multiple categories such as shopping, traveling, etc, and requires both temporal reasoning and textual reasoning to answer questions. Tom et al. \cite{roadtextvqa_icdar_2023}, proposed a dataset for the task of video question answering in the context of driver assistance on road videos. Additionally, a competition \footnote{\footnotesize\url{https://tianchi.aliyun.com/competition/entrance/532050/information}} centered on the task of answering questions based on video content using text was introduced.
% In \footnote{\footnotesize\url{https://tianchi.aliyun.com/competition/entrance/532050/information}}, authors have introduced a competition on text-based video question answering task. 

\begin{table}[h]
\centering
\caption{Analysis of 100 random QA pairs from M4-ViteVQA and NewsVideoQA datasets. }
\resizebox{\columnwidth}{!}{%
\begin{tabular}{@{} l c c @{}} \toprule
{Category} & {M4-ViteVQA (\%)} & {NewsVideoQA (\%)} \\
\midrule
Single Frame & 92.0 & 95.0 \\
Multi Frame & 8.0 & 5.0 \\
Visual Info & 33.0 & 6.0 \\
Textual Info & 95.0 & 100.0 \\
Frame crowded with text & 18.0  & 64.0 \\
Extractive-based & 81.0  & 98.0 \\
Reasoning-based & 5.0  & 2.0 \\
Knowledge-based & 1.0  & 0.0 \\
\bottomrule
\end{tabular}%
}
\label{tab:meta_data_table}
\end{table}

\begin{table*}[t]
    \centering
    \caption{ Performance comparison of different methods on the validation set of M4-ViteVQA dataset. It can be seen that a simple text-only model achieves comparable results and beats the scores of the original multimodal method. }
    % \resizebox{\columnwidth}{!}{%
    \begin{tabular}{@{}lc cccccc@{}} \toprule
    \multirow{2}{4em}{Experiment} & \multirow{2}{4em}{Finetuning} & \multicolumn{2}{c}{Task1 Split1} & \multicolumn{2}{c}{Task1 Split2} & \multicolumn{2}{c}{Task2}  \\
     \cmidrule(l){3-4}\cmidrule(l){5-6}\cmidrule(l){7-8}
     &  & ACC. & ANLS & ACC. & ANLS & ACC. & ANLS  \\
    \midrule
    % Upper Bound & - & 68.44 & 71.0 & 67.22 & 70.2 &  66.40 & 69.8 \\
    T5-ViteVQA & $\checkmark$ & \textbf{23.17} & 30.10 & \textbf{17.59} & 23.10 & 12.30 & 16.10 \\  
    BERT-QA & $\times$ & 9.03 & 17.05 & 8.17 & 15.81 & 10.89 & 18.41 \\
    BERT-QA & $\checkmark$ & 21.96 & \textbf{32.18} & 17.10 & \textbf{26.05} & \textbf{16.01} & \textbf{24.08} \\
    \bottomrule
    \end{tabular}
%     %
% }
    \label{tab:vitevqa_main_exps}
\end{table*}

In this work, we explore the task of text-based video question answering. Firstly, we study and analyze two recently introduced datasets, namely NewsVideoQA \cite{newsvideoqa_wacv_2023} and M4-ViteVQA \cite{vitevqa_neurips_2022}, which includes various types of videos, such as news videos, vlogging, traveling and shopping. We conduct an exploratory analysis to examine the level of visual understanding and multi-frame comprehension required for answering the questions in both datasets. Additionally, we conduct experiments using BERT-QA \cite{bert_naacl_2019}, a text-only model, and demonstrate its effectiveness by achieving comparable results to the original methods that consider both visual and textual information. We also analyze domain adaptation by training on M4-ViteVQA and testing on NewsVideoQA, and vice-versa, revealing insights into cross-domain understanding challenges. 

% By addressing these research aspects, our work offers novel insights and advancements in the realm of video question answering, paving the way for improved understanding and analysis of textual content in videos.

\section{Benchmarking and Experiments}
In this section, we present details of the exploratory analysis and the experiments we conduct.
BERT-QA is a transformer-based encoder-only model pre-trained on a large corpus and further finetuned on SQuAD dataset \cite{squad_emnlp_2016} for question answering (Extractive QA). Extractive QA is the task of extracting a short snippet from the document/context on which the question is asked. The answer `span' is determined by its start and end tokens. It is selected for its effective extractive QA performance, implementation ease, and finetuning, despite limitations like no answer generation or handling yes/no questions. Its ability in extracting answers from textual content makes it a suitable choice for tasks where answers are primarily found in the text of the video. To convert both M4-ViteVQA and NewsVideoQA datasets in SQuAD format, we find the first substring of the answer in the context, which is an approximation of the answer span as followed in \cite{docvqa_wacv_2021}.

%------------------------

\subsection{Exploratory Analysis}

For exploratory analysis, we randomly sample 100 QA pairs from both M4-ViteVQA and NewsVideoQA. For each QA pair, we check the following aspects: i) if the question can be answered by a single frame or need multi-frame information, ii) if the question needs visual information and/or textual information to obtain the answer, iii) if the frame which is essential to obtain the answer, is crowded with text (approximately more than 15 OCR tokens). From Table. \ref{tab:meta_data_table}, it can be seen that for both datasets, information from a single frame is sufficient to obtain answers, which is counter-intuitive to the video question-answering task. From Table. \ref{tab:meta_data_table}, it can also be seen that most of the questions in both datasets need textual information to obtain answers. As M4-ViteVQA contains videos from multiple categories, it contains more questions of visual type compared to NewsVideoQA that contains only news videos. Since both datasets are designed for questions that require reading text to answer questions, this has resulted in minimal questions that require multimodal information. We also check for the answer type: i) extractive, ii) reasoning based, and iii) knowledge-based, and combinations of each type. From Table. \ref{tab:meta_data_table}, it can be seen that most of the questions are extractive in nature and have fewer reasoning-based and knowledge-based questions. However, having more reasoning/knowledge-based questions is crucial, thereby creating the need for better methods beyond the scope of text-only models.

% It was observed that in \cite{vitevqa_neurips_2022}, 81 \% of questions were extractive, 5\% were reasoning based, 1\% were knowledge-based, 2\% were reasoning and knowledge-based, 9\% were reasoning and extractive, and 2\% were extractive and knowledge-based. In \cite{newsvideoqa_wacv_2023}, 98\% questions were extractive and 2\% were reasoning based questions. As it can be observed that nature of the questions is mostly extractive which makes models like BERT-QA a good pick. 

\subsection{BERT-QA experiments}

\textbf{M4-ViteVQA \cite{vitevqa_neurips_2022}:} The M4-ViteVQA dataset consists of two tasks. The first task is divided into two splits and both splits contain evenly distributed question-answer pairs from all video categories in train-val-test sets. In the second task, the training set comprises videos from seven categories, while the question-answer pairs and videos in validation in test splits are exclusively sourced from the remaining two categories. Zhao et el. \cite{vitevqa_neurips_2022} also propose a multimodal video question-answering method: T5-ViteVQA, that combines information from multiple modalities including OCR features, question features, and video features. 

In our experiments on the BERT-QA model, we first sample frames at 1fps and order the OCR tokens of the frames to the default reading order based on the position of the top-left corner of the OCR token. We further concatenate the ordered OCR tokens which becomes the context of the BERT-QA model.
%  To train the BERT-QA model, we order the OCR tokens in each frame sampled at 1fps in default reading order to form the context of the BERT-QA model. sampling 1fps obtain OCR tokens of the 1fps sampled frames which are provided along with the dataset. We concatenate the OCR tokens from the sampled frames to form a context. This context, consisting of OCR tokens from the sampled frames along with the question is considered as the input for the BERT-QA model during training. The model is trained to identify the span of the answer. 
After the training phase, we conduct two types of testing to evaluate the performance of the BERT-QA model. For the first type, we evaluate the model on the entire validation set without checking if the answer is present in the context. This experiment allows us to assess the model's overall ability to obtain answers. In the second type of testing, we specifically focus on questions that have answers in the context.

% -------------------------------------------------

\begin{table}[]
    \centering
    \caption{ Performance comparison of BERT-QA model on M4-ViteVQA dataset when the answer to questions is present in the concatenated list of OCR tokens from evenly sampled frames.}
    \begin{tabular}{@{}lccc@{}} \toprule
     Split &  Acc. &  ANLS  &  No. of QA pairs\\ \midrule
     % Task 1 Split 1  & 19.20 & 25.25 & $\times$ & 911\\
      Task 1 Split 1  &  47.42 &  55.14  &  911\\ 
     % Task 1 Split 2  & 20.30 & 27.19 & $\times$ & 532\\
      Task 1 Split 2  &  42.29 &  48.90  &  532\\ 
     % Task 2  & 25.78 & 30.48 & $\times$ & 318\\
      Task 2  &  38.05 &  43.82  &  318\\
    \bottomrule
    \end{tabular}
    \label{tab:vitevqa_ap_exps}
\end{table}

\textbf{NewsVideoQA \cite{newsvideoqa_wacv_2023}:} This dataset proposes questions on news videos. The dataset has timestamps for each question indicating the frame at which the question was defined. This work also proposes a repurposed baseline: OCR-aware SINGULARITY, which was originally inspired by SINGULARITY \cite{singularity_acl_2023}. OCR-aware SINGULARITY is a multimodal transformer-based video question-answering model that combines information from OCR tokens, questions, and visual information from a randomly sampled frame. 

In this work, we conduct two types of training on this dataset. In the first approach, we train the BERT-QA model using the OCR tokens of the single frame on which the question was defined (BERT-QA-SF: BERT-QA Single Frame). In the second approach, we concatenate the OCR tokens from frames sampled at 1fps which forms the context of the BERT-QA model. (BERT-QA-MF: Multi-frame). By conducting training in both single-frame (BERT-QA-SF) and multi-frame (BERT-QA-MF) setups, we aim to explore the impact of variations in the length of context on the performance of the BERT-QA model. These two training approaches provide insights into the model's ability to obtain answers based on either a specific frame or a broader contextual understanding derived from multiple frames.

% The dataset offers OCR annotations at 2 fps from videos. These are obtained using GoogleOCR, which gives OCR annotations in a default reading order that is beneficial for our experiments (unlike annotations from M4-ViteVQA).  we utilize timestamp annotation of the question to identify the frame it is defined on and obtain OCR tokens with respect to that frame. With this information, we train the BERT-QA model in a single-frame setup, where the context for each question consists of OCR tokens from the corresponding frame. The BERT-QA model is then trained using the multiple-frame setup, where the context encompasses the concatenated OCR tokens from the sampled frames. These two training approaches provide insights into the model's ability to obtain answers based on either a specific frame or a broader contextual understanding derived from multiple frames. We conduct experiments to explore its generalization capabilities and knowledge transfer.

\subsection{Domain Adaptation Experiments}
We conduct experiments to determine if the BERT-QA model can perform or generalize well with the out-of-domain context.  This evaluation aims to determine if the model can provide accurate answers even in unfamiliar video categories and their corresponding contexts. To achieve this understanding, we perform several experiments. We check for the performance of the BERT-QA model trained on the \textbf{Source dataset} followed by testing on the \textbf{Target dataset}. We do this in two settings: i) without finetuning on the target dataset, and ii) with finetuning on the target dataset (Example: Train on NewsVideoQA and test on M4-ViteVQA in two settings i.e. with/without finetuning and vice-versa). By doing these, we try to examine the impact of domain shift and the importance of training the model on videos from diverse categories, where scene text serves as the textual content in one dataset that is M4-ViteVQA, as opposed to embedded text in NewsVideoQA. These experiments help us determine the model's ability to generalize and adapt to the specific categories of videos.

% This evaluation involves training the model using OCR tokens sampled at 1 fps from NewsVideoQA and subsequently finetuning it on the M4-ViteVQA dataset. Firstly, we test the performance of the BERT-QA model trained on the NewsVideoQA: BERT-QA-MF (BERT-QA trained in a multi-frame setup) on the M4-ViteVQA. In the second experiment, we assess the performance of the BERT-QA model trained on the M4-ViteVQA on the NewsVideoQA. We then finetune this model on the NewsVideoQA to evaluate the importance of finetuning a model initially trained on videos from nine different categories. We conduct experiments to assess the BERT-QA model's ability to answer questions across different domains.

\begin{table}[]
    \centering
    \caption{ Performance comparison (Acc.) of M4C, T5-ViteVQA, and BERT-QA on the validation set of {Task 1 Split 1}.}
    \begin{tabular}{@{}lccc@{}} \toprule
     Set &  M4C \cite{m4c_cvpr_2020} &  T5-ViteVQA \cite{vitevqa_neurips_2022} &  BERT-QA \cite{bert_naacl_2019} \\ \midrule
      Easy &  19.30 &  25.09 &  \textbf{25.49}\\
      Hard  &  9.02 &  14.26 &  \textbf{16.34}\\ 
      Text  &  17.26 &  23.08 &  \textbf{31.01}\\
      Vision  &  18.36 &  \textbf{24.21} &  18.82\\  
    \bottomrule
    \end{tabular}
    \label{tab:vitevqa_detailed_exps}
\end{table}

\subsection{Evaluation Metrics and Experimental Setup}
Frequently used in the majority of the works on scene-text based visual and video question answering, we use two evaluation metrics --- Accuracy (Acc.) and Average Normalized Levenshtein Similarity (ANLS). Accuracy is the percentage of questions for which correct answers and predicted answers match exactly. Whereas ANLS is a similarity-based metric that acts softly on minor answer mismatches. More details can be found in \cite{stvqa_iccv_2019}. For all experiments, we train BERT-QA \texttt{bert-large-uncased-\\whole-word-masking-finetuned-squad} for 15 epochs with a batch-size of 16 on 4 GPUs with a learning rate of 2e-05.

\subsection{Quantitative results}

In this section, we present the results and analysis of different experiments. In Table. \ref{tab:vitevqa_main_exps}, we show the results of the performance of T5-ViteVQA and BERT-QA on different tasks and splits on the validation set of the M4-ViteVQA dataset. It can be seen that a simple text-only model achieves comparable results and beats the scores of T5-ViteVQA for certain splits. The results indicate that we need more datasets that require information from multiple modalities and multiple frames which is a concerning limitation in the current datasets. 
% For task 2, even though the model has not seen questions from two categories which are shopping and talking, it still performs better than T5-ViteVQA. This indicates that the current VideoQA datasets based on text need better formulation in terms of creating datasets not being purely text-based.
It can be seen that the BERT-QA relies purely on the OCR output to infer and extract the answer. Therefore, if the OCR output is noisy or if the tokens are incorrectly ordered (errors in default reading order) the model might fail to find the right answer. However, since the ANLS metric acts softly on OCR errors, BERT-QA outperforms T5-ViteVQA on the ANLS metric. 
In Table. \ref{tab:vitevqa_ap_exps}, we show the performance of BERT-QA for the questions that contain answers in the context. We create this test set by checking if the answer is a substring of context. For each of the splits, nearly half of the original questions in the validation set have answers in the context.  In Table. \ref{tab:vitevqa_detailed_exps}, we show the performance comparison---in terms of Accuracy---of two methods: i) M4C \cite{m4c_cvpr_2020}: It uses a multimodal transformer and an iterative answer prediction module. The model answers questions based on scene-text questions on a single image. ii) T5-ViteVQA: method proposed as a baseline in \cite{vitevqa_neurips_2022}, with BERT-QA on the validation set of Task 1 Split 1. It can be seen that BERT-QA outperforms M4C and T5-ViteVQA on different sets. Here, the ``sets" correspond to the type of questions which is provided with the dataset. These sets are: i) easy - answering requires information from a single frame, ii) hard - answering requires information from multiple frames, iii) text - answering requires only reading text, and iv) vision - answering requires both visual and textual information. Only for questions that require visual information, BERT-QA underperforms, yet still manages to obtain decent performance.

% From both the tables, it can be inferred that the quality of the OCR and the default reading order is necessary for BERT-QA to perform well as observed by looking at the ANLS scores. BERT-QA outperforms compared to other baselines on almost all of the sets of questions except for vision. The easy questions account for the questions that require information from a single frame to obtain the answer, whereas hard questions require information from multiple frames to obtain the answer.

\begin{table}[]
    \centering
    \caption{We show the performance of OCR-aware SINGULARITY \cite{newsvideoqa_wacv_2023} and BERT-QA in different settings. BERT-QA-SF: single frame setup, BERT-QA-MF: multi-frame setup on NewsVideoQA.  In the second column, we explain the type of testing. ``12 random frames": considers visual and textual information from 12 random frames, ``single random frame": OCR tokens of a random frame, ``single correct frame": OCR tokens of the correct frame, ``1 frame per second": OCR tokens of frames sampled at 1fps.}
    \resizebox{\columnwidth}{!}{%
    \begin{tabular}{@{}l l  l r@{}} \toprule
     Baseline & Type of testing & Acc. & ANLS  \\ 
      \midrule
      OCR-aware SINGULARITY &  12 random frames & 32.47 & 35.56 \\
      BERT-QA-SF  &  single random frame &  23.71 & 29.47 \\
     % \small BERT-QA  & \footnotesize single correct frame & $\times$ & 33.29 & 43.43 \\
      BERT-QA-SF  &  single correct frame & 46.55  & 56.81 \\
     % \small BERT-QA & \footnotesize all frames & $\times$ & 31.31  & 40.60\\ 
      BERT-QA-MF  &  1 frame per second &  \textbf{52.29} & \textbf{61.12} \\ 
      \bottomrule
    \end{tabular}
    }
    
    \label{tab:newsvideoqa_main_exps}
\end{table}

\begin{table}[!ht]
    \centering
    \caption{Out-of-domain training performance for NewsVideoQA and M4-ViteVQA datasets. ``Source dataset" corresponds to the dataset on which we train the model, and ``Target dataset" corresponds to the dataset we test the model on.}
    \resizebox{\columnwidth}{!}{%
    \begin{tabular}{@{}lccccr@{}}
    \toprule
     Source dataset &  Target dataset &  Finetuning on target  &  Acc. &  ANLS  \\ 
    \midrule
     NewsVideoQA & NewsVideoQA & \checkmark & 52.29 & 61.12 \\
     M4-ViteVQA & NewsVideoQA &  $\times$  & 40.39 & 51.86 \\ 
     M4-ViteVQA &  NewsVideoQA &  \checkmark & 50.41 & 61.04 \\ \midrule
     M4-ViteVQA & M4-ViteVQA & \checkmark & 21.96 & 32.18 \\
     NewsVideoQA & M4-ViteVQA & $\times$  & 7.86 & 12.68 \\
     NewsVideoQA &  M4-ViteVQA & \checkmark & 22.17  & 31.95 \\
    \bottomrule
    \end{tabular}%
    }
    \label{tab:cross_domain_exps}
\end{table}

% \begin{table}[!ht]
%     \centering
%     \caption{Out-of-domain training performance for NewsVideoQA and M4-ViteVQA datasets.}
%     \resizebox{\columnwidth}{!}{%
%     \begin{tabular}{@{}lccccc@{}}
%     \toprule
%     \footnotesize Training & \footnotesize Finetuning & \footnotesize Testing  & \footnotesize Acc. & \footnotesize ANLS  \\ 
%     \midrule
%     \footnotesize M4-ViteVQA & - & \footnotesize NewsVideoQA  & 40.39 & 51.86 \\ 
%     \footnotesize M4-ViteVQA & \footnotesize NewsVideoQA & \footnotesize NewsVideoQA & 50.41 & 61.04 \\
%     \footnotesize NewsVideoQA & - & \footnotesize M4-ViteVQA & 7.86 & 12.68 \\
%     \footnotesize NewsVideoQA & \footnotesize M4-ViteVQA & \footnotesize M4-ViteVQA & 22.17  & 31.95 \\
%     \bottomrule
%     \end{tabular}%
%     }
%     \label{tab:cross_domain_exps}
% \end{table}

In Table. \ref{tab:newsvideoqa_main_exps}, we show results of the performance of different methods on the test set of NewsVideoQA\cite{newsvideoqa_wacv_2023} dataset. OCR-aware SINGULARITY is a model trained in a single-frame setup and is tested on a multi-frame setup (by combining visual and textual information from 12 frames - more details in \cite{newsvideoqa_wacv_2023}). This is followed by results of BERT-QA-SF i.e. trained on OCR context from a single frame and tested by picking a random frame. In the third row, we show the results of BERT-QA when tested with OCR tokens of the frame on which the question was defined (correct frame). In the fourth row, BERT-QA-MF: BERT-QA is trained and tested on a multi-frame setup. In Table. \ref{tab:cross_domain_exps}, we show the results of out-of-domain training performance on both \cite{vitevqa_neurips_2022, newsvideoqa_wacv_2023} datasets.  It can be seen that testing a model initially trained on M4-ViteVQA (Source dataset) achieves decent performance on an out-of-domain NewsVideoQA (target dataset) and vice-versa. By further finetuning on the target dataset, the performance of the model increases. This indicates that the BERT-QA model can effectively generalize across domains through out-of-domain training. More details are present in the supplementary.

% We can see that, for NewsVideoQA has an accuracy of 40 \% by directly testing model trained on M4-ViteVQA. Even though by directly testing M4-ViteVQA trained on NewsVideoQA does not yield enough scores, it achieves better performance by further finetuning the model on M4-ViteVQA. 

\section{Conclusion}

This paper focused on the important task of understanding textual information within videos for question-answering. The study provides insights that current text-based VideoQA datasets majorly focus on extractive answers and provide insights that the degree of visual understanding and multi-frame comprehension in current datasets is limited for better VideoQA using text in videos. Additionally, the paper demonstrates the effectiveness of BERT-QA, a text-only model, in achieving comparable performance to original methods on both datasets and also looks into the domain transfer aspect, by comparing the performances by training on one type of dataset and testing on the other. In future developments, we hope to see datasets that prioritize non-extractive answers and incorporate multimodal questions based on multiple frames to facilitate improved multimodal learning.

\textbf{Acknowledgements.}
This work is supported by MeitY, Government of India.

{\small
\bibliographystyle{ieee_fullname}
\bibliography{egbib}
}

\end{document}

%%%%%%%%% TITLE
\title{Supplementary: Understanding Video Scenes through Text: Insights from Text-based Video Question Answering}

\author{\quad Soumya Jahagirdar$^{1}$  \qquad Minesh Mathew$^{2}$ \qquad Dimosthenis Karatzas$^{3}$ \qquad C. V. Jawahar$^{1}$ \\
{\tt\footnotesize soumya.jahagirdar@research.iiit.ac.in} \enspace \enspace {\tt\footnotesize minesh@wadhwaniai.org} \enspace \enspace {\tt\footnotesize dimos@cvc.uab.es}  \enspace\enspace {\tt\footnotesize jawahar@iiit.ac.in} \\
$^{1}$ CVIT, IIIT Hyderabad, India \enspace\enspace $^{2}$ Wadhwani AI \enspace\enspace $^{3}$ Computer Vision Center, UAB, Spain} 

% \and

\maketitle
\ificcvfinal\thispagestyle{empty}\fi

As discussed in the main paper, we add extra annotations for each question-answer pair a type for M4-ViteVQA \cite{vitevqa_neurips_2022} dataset. These are as follows:

\begin{enumerate}
    \item Extractive-based: These are the type of answers which can be directly extracted from the context which in this case is OCR tokens. (Meaning that the answer is a substring of the concatenated string of OCR tokens)
    \item Reasoning-based: For this type, the answer can be obtained by reasoning over the content in the image (both visual and textual). The examples could be yes/no type of questions where answer is not always present in the text or visual content but it needs the model to reason over multiple modalities.
    \item Knowledge-based: In this type, the answers usually require external knowledge such as knowing that a particular brand name is for a specific product.
\end{enumerate}

\section{Quantitative results}

Tab. \ref{tab:vitevqa_exps} presents the results of the BERT-QA \cite{bert_naacl_2019} model on M4-ViteVQA \cite{vitevqa_neurips_2022} dataset. The evaluation consists of a comprehensive series of experiments. We divide the experiments into two types. In the first type, we consider all the questions from the validation set for all the original splits i.e. \texttt{Task1Split1}, \texttt{Task1Split2}, and \texttt{Task2}. For the second type, we narrow down our analysis to only include question-answer pairs where the answer is found within the context. In both Tab. \ref{tab:vitevqa_exps} and \ref{tab:cross_domain_exps}, it can be seen that we only the results on BERT-QA model for the cases when for the cases where answer is present in the context (for the entire validation set, the results are already shown and compared to T5-ViteVQA in the main paper) because the code for T5-ViteVQA or the pretrained models are not open sourced.

\begin{figure} 
    \centering
    \includegraphics[width=1\linewidth]{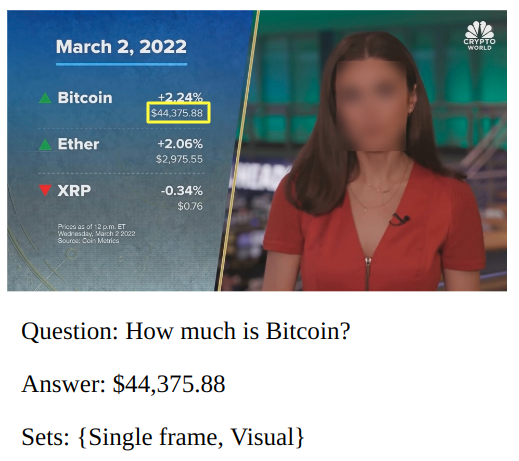}
    \caption{A specific example from M4-ViteVQA \cite{vitevqa_neurips_2022} dataset. For this example, the dataset provides two annotations, (1) whether the question can be answered by a single frame, and (2) whether a question needs visual information along with textual information to obtain answer.
    }
    \label{fig:sets_vis_sf}
\end{figure}

Specifically, we extract the concatenated list of OCR tokens from frames sampled at 1fps and consider only those pairs where the answer appears in this context. It can be seen that with finetuning the performance of BERT-QA increases over all tasks and splits. It can also be observed that the ANLS metric is significantly better than accuracy. This is due to the fact that the OCR tokens do not have ground truth default reading order which is very essential for extractive question-answering models like BERT-QA. It is also because the OCR annotations provided in M4-ViteVQA \cite{vitevqa_neurips_2022} dataset are obtained from open-sourced OCR detection and recognition, which can be improved by using a commercial OCR such as GoogleOCR. In the second type of experiment, where we evaluate only the questions that have answers in the context, we observe improved scores. This outcome is somewhat expected, given that we are specifically selecting question-answer pairs where the answer is present in the context. Consequently, the model's performance is naturally better in this scenario.

\begin{table*}[]
    \centering
    \caption{Performance comparison of BERT-QA model on M4-ViteVQA \cite{vitevqa_neurips_2022} dataset when the answer to questions is present in the concatenated list of OCR tokens from evenly sampled frames.}
    \begin{tabular}{@{}lccccc@{}} \toprule
    Split & Answer present in context & Finetuning & Acc. & ANLS   & No. of QA pairs\\ \midrule
     Task 1 Split 1  & No & $\times$ & 9.03 & 17.05  & 1971\\
     Task 1 Split 1  & No & $\checkmark$ & 21.96 & 32.18  & 1971\\
     Task 1 Split 1  & Yes & $\times$ & 19.20 & 25.25  & 911\\
     Task 1 Split 1  & Yes & $\checkmark$ & 47.42 & 55.14  & 911\\ \midrule

     Task 1 Split 2  & No & $\times$ & 8.17 & 15.81  & 1321\\
     Task 1 Split 2  & No & $\checkmark$ & 17.10 & 26.05  & 1321\\
     Task 1 Split 2  & Yes & $\times$ & 20.30 & 27.19  & 532\\
     Task 1 Split 2  & Yes & $\checkmark$ & 42.29 & 48.90  & 532\\ \midrule

     Task 2  & No & $\times$ & 10.89 & 18.41  & 762\\
     Task 2  & No & $\checkmark$ & 16.01 & 24.08  & 762\\
     Task 2  & Yes & $\times$ & 25.78 & 30.48  & 318\\
     Task 2  & Yes & $\checkmark$ & 38.05 & 43.82  & 318\\ 
    \bottomrule
    \end{tabular}
    \label{tab:vitevqa_exps}
\end{table*}
\begin{table*}[]
    \centering
    \caption{In this table, we show the results of the performance of the BERT-QA model on the test set of the NewsVideoQA \cite{newsvideoqa_wacv_2023} dataset. For the random frame, we sample a frame randomly and consider its OCR tokens as context to the model.}
    \begin{tabular}{@{}l c c c c@{}} \toprule
     Training data & Testing data & Ft & Acc. & ANLS  \\ 
      \midrule
       - & single random frame & $\times$ & 16.78 & 22.47 \\ 
       single correct frame & single random frame & $\checkmark$ & 23.71 & 29.47 \\
       - & single correct frame & $\times$ & 33.29 & 43.43 \\
       single correct frame & single correct frame & $\checkmark$ & 46.55  & 56.81 \\ 
       - & 1fps-sampled-frame & $\times$ & 31.31 & 40.60 \\
      single correct frame & 1fps-sampled-frame & $\checkmark$ & 51.25 & 62.67 \\
       1fps-sampled-frame & single random frame & $\checkmark$ & 17.41 & 20.36 \\
       1fps-sampled-frame & single correct frame & $\checkmark$ & 37.26 & 42.26 \\
      1fps-sampled-frame & 1fps-sampled-frame & $\checkmark$ & 52.29 & 61.12 \\ 
      \bottomrule
    \end{tabular}
    \label{tab:newsvideoqa_exps}
\end{table*}

\begin{table*}[]
    \centering
    \caption{In this table, we show the results of the performance of the BERT-QA model out of domain data.}
    \begin{tabular}{@{}l c c c c@{}} \toprule
      Type of training & Continued finetuning & Testing data & Acc. & ANLS  \\
      \midrule
        M4-ViteVQA Task 1 Split 1 & - & NewsVideoQA & 40.39 & 51.86 \\
        M4-ViteVQA Task 1 Split 1 & NewsVideoQA & NewsVideoQA & 50.41 & 61.04 \\
        M4-ViteVQA Task 1 Split 2 & - & NewsVideoQA & 37.26 & 47.36 \\
        M4-ViteVQA Task 1 Split 2 & NewsVideoQA & NewsVideoQA & 52.81 & 63.54 \\
        M4-ViteVQA Task 2 & - & NewsVideoQA & 34.96 & 46.98 \\
        M4-ViteVQA Task 2 & NewsVideoQA & NewsVideoQA & 53.44 & 64.27 \\
        NewsVideoQA & - & M4-ViteVQA Task 1 Split 1 & 7.86 & 12.68 \\
        NewsVideoQA & M4-ViteVQA Task 1 Split 1 & M4-ViteVQA Task 1 Split 1 & 22.17 & 31.95 \\
        NewsVideoQA & - & M4-ViteVQA Task 2 & 7.34 & 12.17 \\
        NewsVideoQA & M4-ViteVQA Task 2 & M4-ViteVQA 2 & 15.74 & 24.13 \\
      \bottomrule
    \end{tabular}
    \label{tab:cross_domain_exps}
\end{table*}

\begin{figure*} 
    \centering
    \includegraphics[width=1\linewidth]{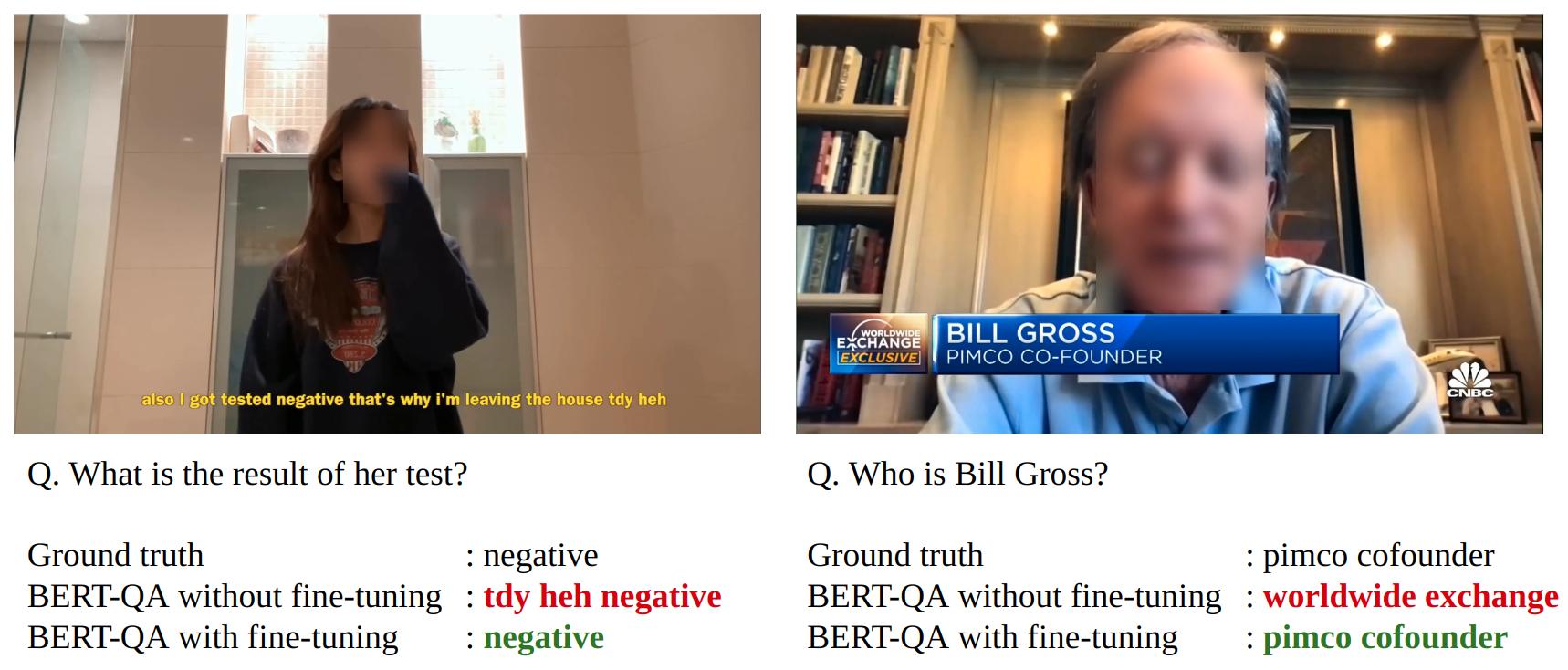}
    \caption{Qualitative results from M4-ViteVQA dataset.
    }
    \label{fig:qual_vitevqa_t1s1_w_wo_ft}
\end{figure*}

In Tab. \ref{tab:newsvideoqa_exps}, we present the results of the BERT-QA model on the NewsVideoQA \cite{newsvideoqa_wacv_2023} dataset. We evaluate the model on two types of training setups: (a) BERT-QA model trained on single frame OCR information. The single frame is obtained from the timestamp of the video where the question was defined. (b) BERT-QA model trained on the concatenated list of OCR tokens from evenly sampled frames per second. In addition to testing types for a single frame (OCR tokens from the frame at which the question was defined) and concatenated lists of OCR tokens (1fps), we also test by considering OCR tokens from a randomly sampled frame. The results indicate that the model tested on OCR tokens from randomly sampled frames performs poorly. This can be attributed to the nature of the BERT-QA model, which is an extractive QA model that predicts the span of the word it considers as the answer from the context. If the context provided to the model is incorrect or unrelated to the question asked, it will extract an incorrect or irrelevant answer span. However, when the correct frame is provided, i.e., the frame at which the question was defined, the model performs better by obtaining the correct answer as the context is correct. In the \texttt{1fps-sampled-frame} experiment, we concatenate OCR tokens from multiple frames (sampled at 1fps), resulting in an increased context. This approach decreases the chances of missing the answer in the context and thus yields improved performance compared to the randomly sampled frames.

In Tab. \ref{tab:cross_domain_exps}, we present the result of the BERT-QA model's ability to generalize by performing out-of-domain finetuning and testing. We first experiment by testing NewsVideoQA dataset's 1fps test set on BERT-QA trained on M4-ViteVQA dataset's \texttt{Task1Split1}. We further finetuning this BERT-QA model trained on M4-ViteVQA \texttt{Task1Split1} on the NewsVideoQA dataset (OCR tokens of 1fps). We repeat the same experiments for the second split i.e. \texttt{Task1Split2}. It can be seen that the performance by testing a model initially trained on M4-ViteVQA i.e. out-of-domain data, achieves decent performance on a dataset that is completely new or out of domain. By further finetuning this model, it achieves even better performance, thereby leveraging training on the M4-ViteVQA dataset. We also experiment by testing \texttt{Task1Split1} on BERT-QA trained on NewsVideoQA. We further finetune it on the M4-ViteVQA dataset. Even in this case, the performance of the finetuned model after initial training on NewsVideoQA achieves better results than by just finetuning BERT-QA directly on M4-ViteVQA for \texttt{Task1Split1}. Therefore it can be inferred that the experiments demonstrate that the BERT-QA model can effectively generalize across domains through out-of-domain finetuning. Additionally, leveraging training on a different dataset improves performance on news datasets showcasing the potential benefits of transfer learning in this task. In conclusion from this section, we can infer that despite the videos being from different domains, it's evident that the model can adapt and assist with the task of extracting text from videos, showcasing its ability to generalize effectively across varied datasets. It not only emphasizes the model's adaptability but also holds the potential to be used in future pretraining techniques, in the context of utilizing combined data from diverse sources, thereby leading to improved overall performance and a deeper understanding of cross-domain information processing.

\section{Qualitative results}

\begin{figure*} 
    \centering
    \includegraphics[width=1\linewidth]{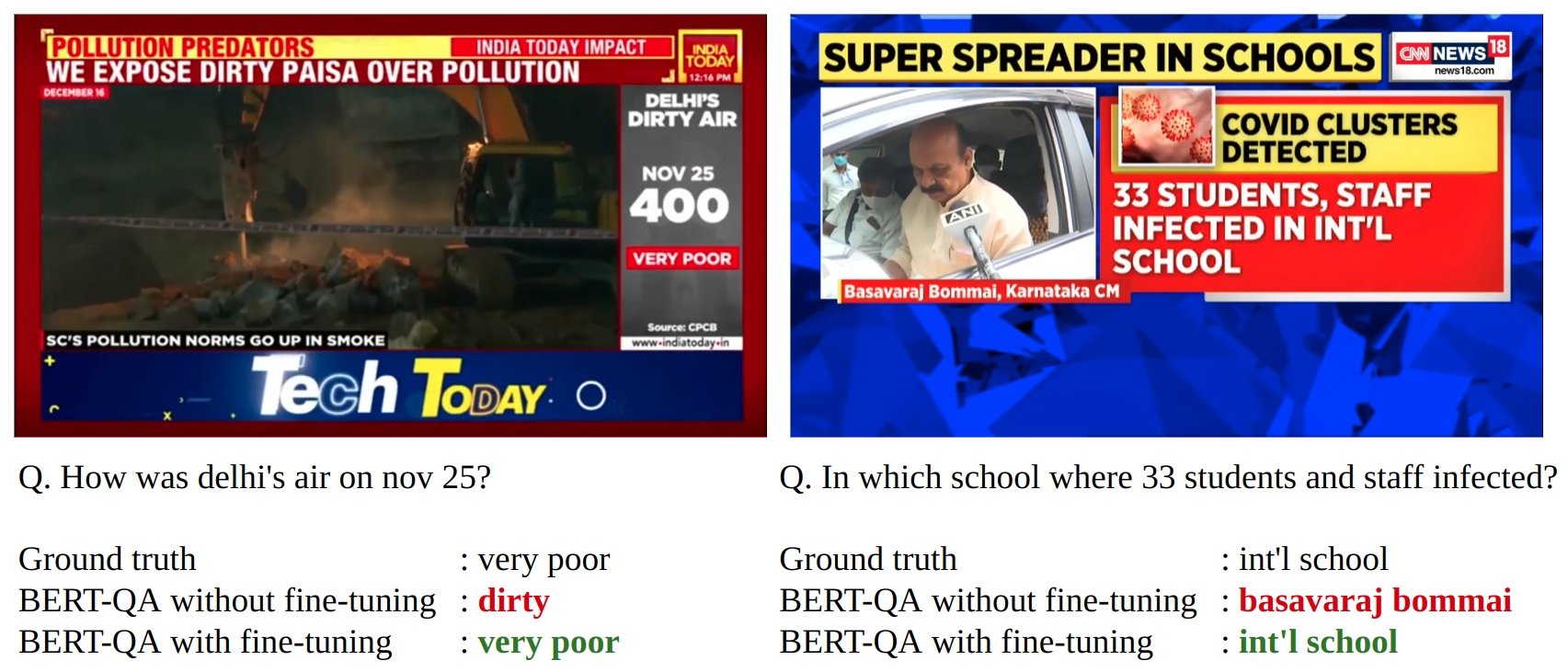}
    \caption{This figure shows the qualitative results of NewsVideoQA \cite{newsvideoqa_wacv_2023} dataset. We show the ground truth, prediction without finetuning the BERT-QA model, and prediction after or with finetuning the BERT-QA model. It can be seen that finetuning helps the model to extract the answers relevant to the questions.
    }
    \label{fig:newsvideoqa_qual_w_wo_ft}
\end{figure*}

\begin{figure*} 
    \centering
    \includegraphics[width=1\linewidth]{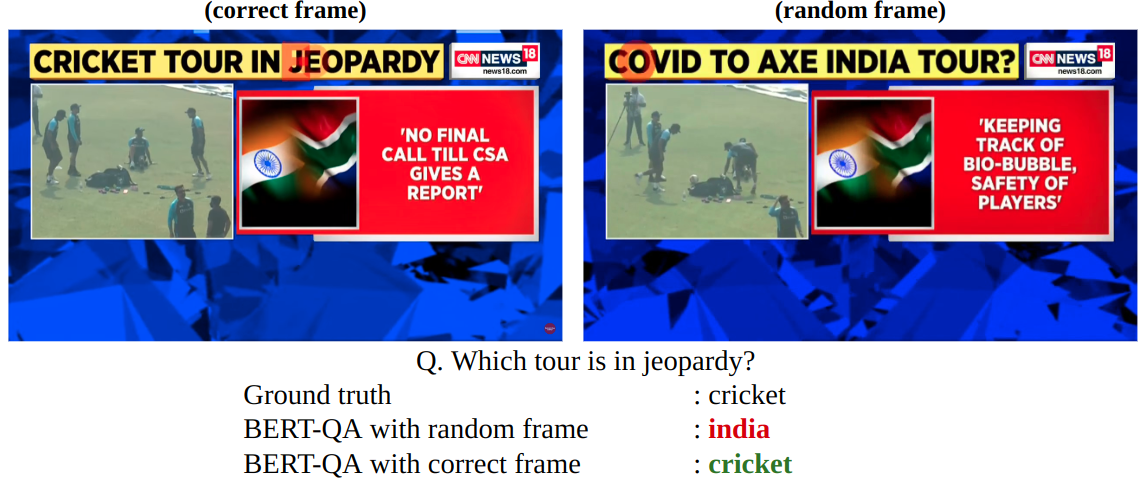}
    \caption{This figure shows the qualitative results of NewsVideoQA \cite{newsvideoqa_wacv_2023} dataset. We show the ground truth, prediction of the BERT-QA model with context as the concatenated list of OCR tokens of the randomly sampled frame, and prediction of the BERT-QA model with context as the concatenated list of OCR tokens of the frame on which the question was defined. It can be seen that the text in the random frame is not sufficient for the model to obtain the answer. Whereas if we give OCR tokens of the frame where the question was defined, it obtains the correct answer.
    }
    \label{fig:newsvideo_correct_random_frame}
\end{figure*}

\begin{figure*} 
    \centering
    \includegraphics[width=1\linewidth]{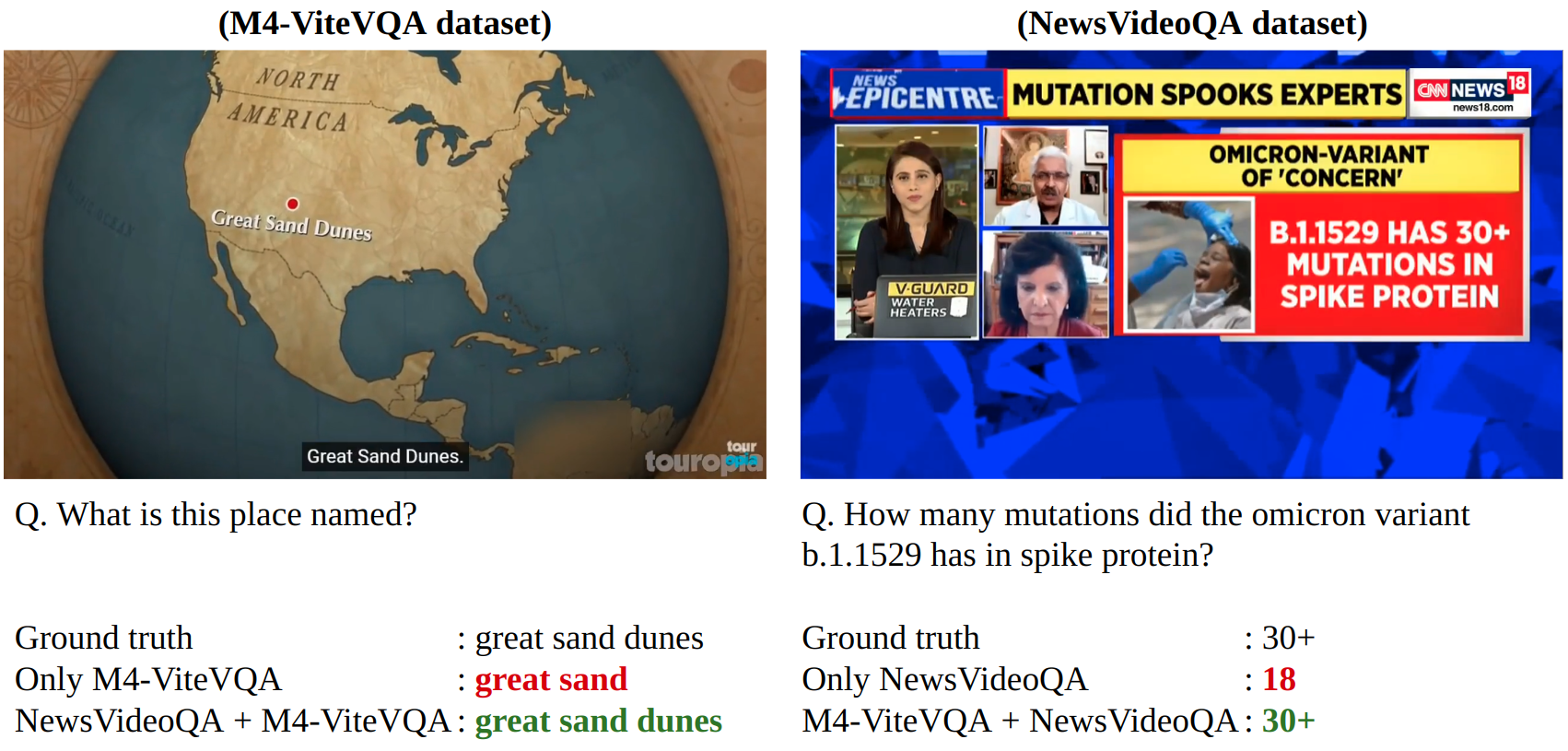}
    \caption{This figure shows the qualitative results for cross-domain training experiments. It can be seen that, when a dataset is trained on its own training set, and tested, the predictions in some cases are not accurate. Whereas out-of-domain training helps the model to understand the extractive nature of the questions thereby increasing the capability to find the correct answer by providing more context.
    }
    \label{fig:cross_domain}
\end{figure*}

In this section, we present qualitative analyses conducted on the two datasets which are NewsVideoQA \cite{newsvideoqa_wacv_2023} and M4-ViteVQA \cite{vitevqa_neurips_2022}, to gain deeper insights. Fig. \ref{fig:newsvideoqa_qual_w_wo_ft} showcases qualitative results obtained from the NewsVideoQA dataset. We compared the ground truth with the predictions made by the BERT-QA model before and after finetuning. The results demonstrate that finetuning helps and improves the model's ability to extract relevant answers related to the questions. Similarly for the M4-ViteVQA dataset, we show the qualitative results in Fig. \ref{fig:qual_vitevqa_t1s1_w_wo_ft}. In Fig. \ref{fig:newsvideo_correct_random_frame}, we present qualitative results from the NewsVideoQA dataset. We compare the predictions of the BERT-QA models using context from OCR tokens from randomly sampled frames and context from the frame on which the question was defined. The results indicate that text in the random frame is insufficient for the model to obtain accurate answers. However, when provided with OCR tokens from the frame where the question was defined, the model successfully obtains the correct answer. In Fig. \ref{fig:cross_domain}, we show the results for the out-of-domain experiments.

% In this section, we provide some qualitative analysis for different experiments and analysis carried out to obtain insights into the two datasets, \cite{vitevqa_neurips_2022} and \cite{newsvideoqa_wacv_2023}. In Fig. \ref{fig:newsvideoqa_qual_w_wo_ft}, we show the qualitative results of NewsVideoQA \cite{newsvideoqa_wacv_2023} dataset. We show the ground truth, prediction without finetuning BERT-QA model, and prediction after or with finetuning BERT-QA model. It can be seen that finetuning helps the model to extract the answers relevant to the questions. In Fig. \ref{fig:newsvideo_correct_random_frame}, we show qualitative results of NewsVideoQA \cite{newsvideoqa_wacv_2023} dataset. We show the ground truth, prediction of BERT-QA model with context as the concatenated list of OCR tokens of the randomly sampled frame, and prediction of BERT-QA model with context as the concatenated list of OCR tokens of the frame on which the question was defined. It can be seen that the text in the random frame is not sufficient for the model to obtain the answer. Whereas if we give OCR tokens of the frame where the question was defined, it obtains the correct answer. In Fig. \ref{fig:qual_vitevqa_t1s1_w_wo_ft} we show the results for \cite{vitevqa_neurips_2022} dataset.

{\small
\bibliographystyle{ieee_fullname}
\bibliography{egbib}
}